\documentclass{article}

\usepackage{amsmath}
\usepackage{amsthm}
\usepackage{amssymb}
\usepackage{mathtools}
\usepackage{bm}

\usepackage{microtype}
\usepackage{graphicx}
\usepackage{booktabs} 

\usepackage{tango-colors}

\usepackage{tikz}
\usepackage{tikz-qtree}
\usetikzlibrary{arrows}
\usetikzlibrary{shapes}

\usepackage{pgfplots}
\pgfplotsset{compat=1.13}
\usepgfplotslibrary{groupplots}

\PassOptionsToPackage{hyphens}{url}
\usepackage{hyperref}
\usepackage{doi}

\usepackage[accepted]{icml2018}

\usepackage[utf8]{inputenc}
\usepackage[T1]{fontenc}
\usepackage{csquotes}

\newtheorem{thm}{Theorem}

\theoremstyle{definition}

\newcommand{\R}{\mathbb{R}}

\newcommand{\mat}[1]{\bm{#1}}
\newcommand{\transp}[1]{{#1}^T}
\newcommand{\effic}{\mathcal{O}}

\newcommand{\grad}[1]{\nabla_{#1}}

\newcommand{\node}{x}
\newcommand{\lnode}{x}
\newcommand{\rnode}{y}

\newcommand{\tree}{\bar x}
\newcommand{\ltree}{\bar x}
\newcommand{\rtree}{\bar y}

\newcommand{\trees}{\mathcal{T}}

\newcommand{\siz}[1]{|#1|}
\newcommand{\childidx}{r}
\newcommand{\childlim}{R}

\newcommand{\alphabet}{\mathcal{X}}
\newcommand{\sym}[1]{\text{\texttt{#1}}}
\newcommand{\gap}{-}

\newcommand{\dataidx}{i}

\newcommand{\rdataidx}{j}
\newcommand{\datalim}{m}
\newcommand{\pospairs}{P}
\newcommand{\negpairs}{N}
\newcommand{\loss}{E}

\newcommand{\margin}{\eta}
\newcommand{\regul}{\beta}
\newcommand{\sparsity}{\lambda}
\newcommand{\simil}{k}
\newcommand{\coeff}{\alpha}
\newcommand{\sign}{\mathrm{sign}}

\newcommand{\alphlim}{U}
\newcommand{\embedmat}{\mat A}
\newcommand{\embeddim}{V}
\newcommand{\embedvec}{\vec a}

\newcommand{\proto}{\bar w}
\newcommand{\cls}{\ell}
\newcommand{\protoidx}{k}
\newcommand{\protolim}{K}
\newcommand{\err}{E}
\newcommand{\nonlin}{\Phi}
\newcommand{\likelihood}{\mathcal{L}}
\newcommand{\relproj}{\Omega}

\newcommand{\edit}{\delta}

\newcommand{\editlim}{T}

\newcommand{\dist}{d}
\newcommand{\cost}{c}
\newcommand{\lnodeidx}{i}

\newcommand{\rnodeidx}{j}
\newcommand{\rnodelim}{n}

\newcommand{\freqmat}{\mat P}

\tikzstyle{point}=[circle, inner sep=0pt, minimum size=3mm, line width=0.5mm, anchor=center]
\tikzstyle{textnode}=[draw=none, fill=none]
\tikzstyle{proto}=[diamond, inner sep=0pt, minimum size=5mm, line width=0.5mm, anchor=center]
\tikzstyle{edge}=[->, >=stealth', shorten <=2pt, shorten >=2pt, auto, line width=0.5mm]
\tikzstyle{class0color}=[aluminium6]
\tikzstyle{class0}=[draw=aluminium6, fill=aluminium4, text=aluminium6]
\tikzstyle{class1color}=[skyblue3]
\tikzstyle{class1}=[draw=skyblue3, fill=skyblue1, text=skyblue3]
\tikzstyle{class2color}=[orange3]
\tikzstyle{class2}=[draw=orange3, fill=orange1, text=orange3]

\begin{document}

\twocolumn[
\icmltitle{Tree Edit Distance Learning via Adaptive Symbol Embeddings}




\begin{icmlauthorlist}
\icmlauthor{Benjamin Paaßen}{bi}
\icmlauthor{Claudio Gallicchio}{pi}
\icmlauthor{Alessio Micheli}{pi}
\icmlauthor{Barbara Hammer}{bi}
\end{icmlauthorlist}

\icmlaffiliation{bi}{Cognitive Interaction Technology, Bielefeld University, Germany}
\icmlaffiliation{pi}{Department of Computer Science, University of Pisa, Italy}
\icmlcorrespondingauthor{Benjamin Paaßen}{bpaassen@techfak.uni-bielefeld.de}

\icmlkeywords{metric learning, trees, structured data, symbol embeddings, learning vector quantization}

\vskip 0.3in
]



\printAffiliationsAndNotice{}  

\begin{abstract}
Metric learning has the aim to improve classification accuracy by learning a distance measure
which brings data points from the same class closer together and pushes data points
from different classes further apart. Recent research has demonstrated that metric learning
approaches can also be applied to trees, such as molecular structures,
abstract syntax trees of computer programs, or syntax trees of natural language, by learning the
cost function of an edit distance, i.e.\ the costs of replacing, deleting, or inserting nodes in a
tree.
However, learning such costs directly may yield an edit distance which violates metric
axioms, is challenging to interpret, and may not generalize well.
In this contribution, we propose a novel metric learning approach for trees which we
call embedding edit distance learning (BEDL) and which learns
an edit distance indirectly by embedding the tree nodes as vectors, such that the Euclidean
distance between those vectors supports class discrimination. We learn such embeddings by
reducing the distance to prototypical trees from the same class and increasing the distance
to prototypical trees from different classes.
In our experiments, we show that BEDL improves upon the
state-of-the-art in metric learning for trees on six benchmark data sets, ranging from
computer science over biomedical data to a natural-language processing data set
containing over 300,000 nodes.
\end{abstract}

\section{Introduction}

Many classification approaches in machine learning explicitly or implicitly rely on some measure
of \emph{distance} \cite{Kulis2013,Bellet2014,Mokbel2015Neurocomputing}. This is particularly
apparent in case of the $k$-nearest neighbor classifier which classifies data points by
assigning the label of the majority of the $k$ \emph{nearest neighbors} according to a
given distance \cite{Cover1967}; or in case of \emph{learning vector quantization}
approaches which classify data points by assigning the label of the closest prototype according
to a given distance \cite{Kohonen1995}. The success of such machine learning approaches
hinges on the distance being \emph{discriminative}, that is, data points from the same
class being generally closer compared to data points from different classes.
If the distance does not fulfill this criterion,
one has to \emph{adapt} or \emph{learn} the distance measure with respect to the data, which is the
topic of \emph{metric learning} \cite{Kulis2013,Bellet2014}.

Most prior research in metric learning has focused on learning a generalization of the Euclidean
distance according to some cost function \cite{Kulis2013,Bellet2014}.
However, the Euclidean distance is not applicable to non-vectorial data, such as protein
sequences, abstract syntax trees of computer programs, or syntax trees of natural language.
To process these kinds of data, \emph{edit distances} are a popular option, in particular the
tree edit distance \cite{Zhang1989}. In this contribution, we develop a novel metric learning
scheme for the tree edit distance which we call embedding edit distance learning (BEDL).

While past research on metric learning for trees does exist \cite{Bellet2014}, BEDL
goes beyond the state-of-the-art in multiple aspects:
\begin{itemize}
\item Based on the work of \citet{Bellet2012}, we provide a generalized re-formulation of the edit
distance which lends itself to metric learning, and can be applied to any kind of edit distance which
uses replacement, deletion, and insertion operations. Furthermore, we consider not only
\emph{one} optimal edit script for metric learning, but \emph{all} co-optimal edit
scripts via a novel forward-backward algorithm.
\item Our approach requires only a linear number of data tuples for metric learning, as we
represent classes by few \emph{prototypes}, which are selected via median learning vector
quantization \cite{Nebel2015}.
\item Most importantly, we do not directly learn the operation costs for the string edit
distance, but instead learn a vectorial embedding of the label alphabet for our data
structures, which yields Euclidean operation costs. This re-formulation ensures
that the resulting edit distance conforms to all metric axioms. Further, we can interpret
the resulting embedding vectors via visualization, their pairwise distances and norms.
\end{itemize}

We begin by discussing related work, then we describe BEDL in more detail, and
finally we evaluate BEDL experimentally and discuss the results.

\section{Related Work}

Our work is related to multiple areas of machine learning, most notably distances on structured
data, metric learning, and vector embeddings.

In the past decades, multiple distance measures for structured data - i.e.\
sequences, trees, and graphs - have been suggested. In particular, one could define a distance based
on existing string, tree, and graph kernels \cite{DaSanMartino2010}, such as Weisfeiler-Lehman Graph
Kernels \cite{Shervashidze2011}, topological distance-based tree kernels \cite{Aiolli2015},
or deep graph kernels \cite{Yanardag2015}. Such kernels achieve state-of-the-art results on
structured data and can be adapted to training data via multiple-kernel
learning \cite{Aiolli2015MKL}, or kernels based on Hidden-Markov-Model states \cite{Bacciu2018}. 
Kernels, however, have drawbacks in terms of interpretability, as a higher
distance value does not necessarily relate to any kind of intuitive difference between the input
trees. Further, kernel matrices are by definition limited to be positive semi-definite,
which may be an undue restriction for certain data sets \cite{Schleif2015}.

If one strives for an interpretable measure of distance, \emph{edit distances} are a popular
choice, for example for the comparison of protein sequences in bioinformatics
\cite{Smith1981}, or abstract syntax trees for intelligent tutoring systems
\cite{Paassen2018JEDM}. Here, we focus on the tree edit distance,
which permits deletions, insertions, and replacements of single nodes to transform an ordered
tree $\ltree$ into another ordered tree $\rtree$ \cite{Zhang1989}. Such ordered trees are the
most general data structures which can still be treated efficiently via edit distances, as
edit distances on unordered trees and general graphs are provably NP-hard \cite{Zhang1992,Zeng2009}.
Furthermore, the tree edit distance includes the edit distance on sequences as a special case,
such that it can be seen as a representative for edit distances as such.

Metric learning for the tree edit distance corresponds to adapting the costs of edit operations
in order to bring trees from the same class closer and push trees from different
classes further apart. Almost all of the existing approaches, however, only bring trees from the
same class closer together \cite{Bellet2014}. For example, \citet{Boyer2007} have proposed to
replace the tree edit distance by the negative log probability of all tree edit scripts which
transform the left input tree $\ltree$ into the right input tree $\rtree$.
Accordingly, the costs of edit operations change to probabilities of replacing, deleting, or inserting
a certain label. These edit probabilities are adapted to maximize the probability that
trees from the same class are edited into each other \cite{Boyer2007}.
To replace generative models by discriminative ones, \citet{Bellet2012,Bellet2016}, 
have proposed to learn an edit distance $\dist$, such that the corresponding similarity
$2 \cdot \exp[-\dist(\ltree, \rtree)] - 1$ is \enquote{good} as defined by the goodness-framework
of \cite{Balcan2008}. Goodness according
to this framework means that a linear separator with low error between the classes exists 
in the space of similarities \cite{Balcan2008,Bellet2012}. \citet{Bellet2012} have experimentally
shown that this approach outperforms generative edit distance metric learning and have also
established  generalization guarantees based on the goodness framework. Therefore, this
\emph{good edit similarity learning} (GESL) approach of \citet{Bellet2012} is our main
reference method.

Our novel approach is strongly inspired by GESL. However, our approach goes beyond GESL in
key aspects. First, we utilize a different cost function, namely the generalized learning
vector quantization (GLVQ) cost function, which quantifies how much closer
every data point is to the closest prototypical example from the same class compared to the
closest prototypical example from another class \cite{Sato1995}. Just as GESL, LVQ methods
are theoretically well-justified because it yields a maximum-margin classifier \cite{Schneider2009},
and have been successfully applied for metric learning on the string edit distance in the past 
\cite{Mokbel2015Neurocomputing,Paassen2016Neurocomputing}.
However, in contrast to GESL, GLVQ also provides a principled way to select prototypical examples
for metric learning, and is flexible enough to not only learn a cost matrix, but also
a vectorial embedding of the tree labels,
such that the Euclidean distance on these embeddings provides a discriminative cost function.

While embedding approaches are common in the literature, prior work has focused mostly on embedding
trees \emph{as a whole}, for example via graph kernel approaches \cite{Aiolli2015,Bacciu2018,DaSanMartino2010,Shervashidze2011,Yanardag2015},
recursive neural networks \cite{Gallicchio2013,Hammer2007,Sentiment}, or dimensionality reduction approaches \cite{Gisbrecht2015}.
In this contribution, we wish to obtain an embedding for the single elements of a tree
and maintain the tree structure.
As of yet, such approaches only exist for sequences, namely in the form of recurrent neural
network for natural language processing tasks \cite{Ch0214,Sutskever2014}.
In addition to word embeddings for trees, our approach also provides a corresponding tree
edit distance, which is optimized for classification, and offers an intuitive view on the
data, supporting applications like intelligent tutoring systems \cite{Paassen2018JEDM}.

\section{Background}

In this section, we revisit the basic problem of tree edit distance learning by first
introducing the tree edit distance of \citet{Zhang1989}, as well as the metric learning
formalization suggested by \citet{Bellet2012}.

\subsection{Tree Edit Distance}

We define a tree $\tree$ over some set $\alphabet$ as $\node(\tree_1, \ldots, \tree_\childlim)$
where $\node \in \alphabet$ and $\tree_1, \ldots, \tree_\childlim$ is a (potentially empty)
list of trees over $\alphabet$. We denote the set of all possible trees over $\alphabet$
as $\trees(\alphabet)$. Further, we call $\node$ the \emph{label} of the tree. 
We define the \emph{size} of a tree $\node(\tree_1, \ldots, \tree_\childlim)$
as $\siz{\tree} := 1 + \sum_{\childidx=1}^\childlim \siz{\tree_\childidx}$.
Finally, we call a list of trees $\tree_1, \ldots, \tree_\childlim$ a \emph{forest}.
Note that every tree is also a forest.

Next, we introduce \emph{edits} over trees. In general, a tree edit $\edit$ is
a function which transforms a forest into a forest \cite{Paassen2018JEDM}.
In this particular case, we are only concerned with three kinds of edits, namely deletions,
which remove a certain label from a forest; insertions, which insert a certain label into a
forest; and replacements which remove a certain label from a forest and put another label in its
place. For example, deleting $\sym{x}$ from a tree $\sym{x}(\sym{y}, \sym{z})$ results in the forest $\sym{y}, \sym{z}$. Inserting
$\sym{x}$ into this forest as parent of $\sym{y}$ results in the forest $\sym{x}(\sym{y}), \sym{z}$. Finally, replacing $\sym{z}$
with $\sym{q}$ in this forest results in the forest $\sym{x}(\sym{y}), \sym{q}$.

We associate each edit with a \emph{cost} via a function $\cost : (\alphabet \cup \{\gap\})^2 \to \R$.
In particular, we define the cost of a deletion of label $\lnode$ as $\cost(\lnode, \gap)$,
the cost of an insertion of label $\rnode$ as $\cost(\gap, \rnode)$,
and the cost of a replacement of label $\lnode$ with label $\rnode$ as $\cost(\lnode, \rnode)$.
We define the cost of a \emph{sequence} of edits $\edit_1, \ldots, \edit_\editlim$ as the
sum over the costs of all edits.

Finally, we define the tree edit distance $\dist_\cost(\ltree, \rtree)$ between any two trees
$\ltree$ and $\rtree$ according to $\cost$ as the cost of the cheapest sequence of edits that
transforms $\ltree$ to $\rtree$. For example, if all edits have a cost of $1$, the edit
distance $\dist_\cost( \sym{x}(\sym{y}, \sym{z}), \sym{q}(\sym{z}(\sym{q})) )$ between the trees $\sym{x}(\sym{y}, \sym{z})$ and $\sym{q}(\sym{z}(\sym{q}))$ is $3$
because the cheapest sequence of edits is to replace $\sym{x}$ with $\sym{q}$, delete $\sym{y}$, and insert $\sym{q}$.

\citet{Zhang1989} showed that the tree edit distance can be computed efficiently using a
dynamic programming algorithm if $\cost$ is a pseudo-metric, meaning that
$\cost$ is a non-negative and symmetric function, such that $\cost(\lnode, \lnode) = 0$ for
any $\lnode \in \alphabet$, and such that the triangular inequality is fulfilled.

\begin{thm}
Let $\cost$ be a pseudo-metric on $\alphabet \cup \{\gap\}$. Then, the
corresponding tree edit distance $\dist_\cost(\ltree, \rtree)$
can be computed in $\effic(\siz{\ltree}^2 \cdot \siz{\rtree}^2)$ 
using a dynamic programming scheme.

Conversely, if $\cost$ violates the triangular inequality, the dynamic
programming scheme overestimates the tree edit distance.

\begin{proof}
Refer to \citet{Zhang1989} for a proof of the first claim, and refer to
the supplementary material \cite{Paassen2018ICMLSupp} for a proof of the second claim.
\end{proof}
\end{thm}

Beyond enabling us to compute the tree edit
distance efficiently, pseudo-metric cost functions $\cost$ also ensure
that the resulting tree edit distance $\dist_\cost$ is a pseudo-metric
itself.

\begin{thm}
Let $\cost$ be a pseudo-metric on $\alphabet \cup \{\gap\}$. Then, the
corresponding tree edit distance $\dist_\cost$ is a pseudo-metric on the
set of possible trees over $\alphabet$.

However, if $\cost$ violates any of the pseudo-metric properties
(except for the triangular inequality), we can construct examples
such that $\dist_\cost$ violates the same pseudo-metric properties.

\begin{proof}
Refer to the supplementary material \cite{Paassen2018ICMLSupp}.
\end{proof}
\end{thm}

Both of these theorems make a pseudo-metric cost function $\cost$
desirable. However, ensuring pseudo-metric properties on $\cost$ may be
challenging in metric learning, which is one of our key motivations
for vectorial embeddings.

\subsection{Tree Edit Distance Learning}
\label{sec:ted_learning}

Tree edit distance learning essentially means to adapt the cost
function $\cost$, such that the resulting tree edit distance
$\dist_\cost$ is better suited for the task at hand.
Following \citet{Bellet2012,Bellet2014}, we frame
tree edit distance learning as minimizing some loss function over a set of \emph{positive} pairs
of trees $\pospairs \subset \trees(\alphabet)^2$ and \emph{negative} pairs of trees
$\negpairs \subset \trees(\alphabet)^2$, that is, trees which should be close and far away
respectively. In particular, given a loss function $\loss$ we wish to solve the optimization
problem:
\begin{equation}
\min_\cost \loss(\dist_\cost, \pospairs, \negpairs)
\end{equation}

In our contribution, we build upon the \emph{good edit similarity learning} (GESL)
approach of \citet{Bellet2012}, who propose the loss function
\begin{align}
\loss(\dist_\cost, \pospairs, \negpairs) = &\regul \cdot \lVert \cost \rVert^2 +
\sum_{(\ltree, \rtree) \in \pospairs} [\dist_\cost(\ltree, \rtree) - \margin]_+  \label{eq:gesl_loss} \\
+ &\sum_{(\ltree, \rtree) \in \negpairs} [\log(2) + \margin - \dist_\cost(\ltree, \rtree)]_+
\notag
\end{align}
where $[ \mu ]_+ = \max \{0, \mu\}$ denotes the hinge loss,
$\margin \in [0, \log(2)]$ is a slack variable permitting higher distances between
positive pairs if negative pairs are further apart, $\regul$ is a scalar
regularization constant, and $\lVert \cost \rVert^2$ denotes
$\sum_{\lnode \in \alphabet \cup \{ \gap \}} \sum_{\rnode \in \alphabet \cup \{ \gap \}} \cost(\lnode, \rnode)^2$.
As positive and negative pairs, \citet{Bellet2012} propose to use the $\protolim$
closest neighbors in the same class and the $\protolim$ furthest data points
from a different class respectively, where \enquote{closeness} refers to
the tree edit distance obtained via some initial, default cost function $\cost_0$
\cite{Bellet2012}.

Note that minimizing the loss function~\ref{eq:gesl_loss} is infeasible because changing
the parameters $\cost$ may change other edit sequences to become the cheapest option
which in turn induces non-differentiable points in the loss function \cite{Mokbel2015Neurocomputing}.
Therefore, \citet{Bellet2012} suggest to compute the cheapest edit scripts according
to a default cost function $\cost_0$ and keep them fixed for the optimization.
More precisely, let $\freqmat_{\cost_0}(\ltree, \rtree)$ be a matrix of size
$\siz{\ltree} + 1 \times \siz{\rtree} + 1$, such that $\freqmat_{\cost_0}(\ltree, \rtree)_{\lnodeidx, \rnodeidx}$
is $1$ if and only if the cheapest edit script that transforms $\ltree$ to
$\rtree$ according to the cost function $\cost_0$ contains a replacement
of node $\lnode_\lnodeidx$ with node $\rnode_\rnodeidx$, where
$\lnode_\lnodeidx$ is the $\lnodeidx$th node in tree $\ltree$ according to
pre-order and $\rnode_\rnodeidx$ is the $\rnodeidx$th node in tree $\rtree$
according to pre-order. Further, if the cheapest edit script deletes $\lnode_\lnodeidx$,
we define $\freqmat_{\cost_0}(\ltree, \rtree)_{\lnodeidx, \siz{\rtree} + 1} = 1$ and
if the cheapest edit script inserts $\rnode_\rnodeidx$, we define
$\freqmat_{\cost_0}(\ltree, \rtree)_{\siz{\ltree} + 1, \rnodeidx} = 1$. We define
all other entries of $\freqmat_{\cost_0}(\ltree, \rtree)$ as zero.
We can compute the matrix $\freqmat_{\cost_0}(\ltree, \rtree)$ in
$\effic(\siz{\ltree}^2 \cdot \siz{\rtree} + \siz{\ltree} \cdot \siz{\rtree}^2)$ via backtracing
(refer to the supplementary material \cite{Paassen2018Arxiv} for details).

Using this
matrix, we can define the \emph{pseudo edit distance} $\tilde \dist_\cost(\ltree, \rtree)$,
which we define as follows.
\begin{equation}
\tilde \dist_\cost(\ltree, \rtree) = \sum_{\lnodeidx = 1}^{\siz{\ltree} + 1} \sum_{\rnodeidx = 1}^{\siz{\rtree} + 1}
\freqmat_{\cost_0}(\ltree, \rtree)_{\lnodeidx, \rnodeidx} \cdot \cost(\lnode_\lnodeidx, \rnode_\rnodeidx) \label{eq:freqmat}
\end{equation}
where we define $\lnode_{\siz{\ltree} + 1} = \rnode_{\siz{\rtree} + 1} := \gap$.
GESL now minimizes the loss function~\ref{eq:gesl_loss} with respect to the pseudo-edit
distance, which is a quadratic optimization problem.

\citet{Bellet2012} show that GESL optimizes the \enquote{goodness} of
the similarity measure $\simil(\ltree, \rtree) = 2 \cdot \exp(-\tilde \dist_\cost(\ltree, \rtree)) - 1$.
The concept of goodness has been introduced by \citet{Balcan2008} and quantifies how
well a given similarity measure lends itself for binary classification.
In particular, assume trees $\tree_1, \ldots, \tree_\datalim$ with class assignments
$\cls(\tree_1), \ldots, \cls(\tree_\datalim)$. Then, we wish to learn
parameters $\vec \coeff \in \R^\datalim$, such that we can classify a new tree $\tree$
via the predictive function $f(\tree) = \sign(\sum_{\dataidx = 1}^\datalim \coeff_\dataidx \cdot \simil(\tree, \tree_\dataidx))$.
We can learn the parameters $\vec \coeff$ by solving the linear minimization problem:
\begin{equation*}
\min_{\vec \coeff} \sum_{\dataidx = 1}^\datalim \Big[ 1 - \cls(\tree_\dataidx) \cdot \sum_{\rdataidx = 1}^\datalim \coeff_{\rdataidx} \cdot \simil(\tree_\dataidx, \tree_{\rdataidx}) \Big]_+
+ \sparsity \cdot \lVert \vec \coeff \rVert_1
\end{equation*}
where $\sparsity$ is a hyper-parameter regulating the L1 regularization, and hence the sparsity,
of $\vec \coeff$.

Recall that GESL optimizes the pseudo edit distance
$\tilde \dist_\cost(\ltree, \rtree)$ instead of the edit distance $\dist_\cost(\ltree, \rtree)$,
and that the theory provided by \citet{Bellet2012} does not guarantee the goodness
of the actual edit distance $\dist_\cost(\ltree, \rtree)$.
Indeed, it may occur that the loss $\loss(\dist_\cost, \pospairs, \negpairs)$
for the actual tree edit distance $\dist_\cost$ is considerably larger than
the loss $\loss(\tilde \dist_\cost, \pospairs, \negpairs)$.

\begin{thm} \label{thm:gesl_degeneration}
There exists combinations of an alphabet $\alphabet$, positive pairs $\pospairs$, negative pairs
$\negpairs$, a default cost function $\cost_0$, and a regularization constant $\regul$, such that
the cost function $\cost_1$ learned by GESL is not a pseudo-metric, and yields a loss 
$\loss(\dist_{\cost_1}, \pospairs, \negpairs) > \loss(\tilde \dist_{\cost_1}, \pospairs, \negpairs)$,
as well as $\loss(\dist_{\cost_1}, \pospairs, \negpairs) > \loss(\dist_{\cost_0}, \pospairs, \negpairs)$.

\begin{proof}
Refer to the supplementary material \cite{Paassen2018ICMLSupp}.
\end{proof}
\end{thm}

Overall, we identify three key points in GESL we would like to address.
First, we would like to select positive and
negative pairs in a principled fashion, in contrast to the ad-hoc scheme
of choosing the closest trees from the same class and the furthest
trees from another class. Second, we would like to enhance the coupling between the pseudo tree edit
distance $\tilde \dist_\cost$ to the actual tree edit distance
$\dist_\cost$. Third, we would like to ensure pseudo-metric properties on $\cost$.

\section{Method}

In this section we introduce a novel method for tree edit distance learning.
We start by selecting positive and negative pairs for metric learning via 
median learning vector quantization. Then, we introduce metric learning
using the generalized learning vector quantization cost function.
Finally, we propose a novel parameterization of the edit cost function
$\cost$ via symbol embeddings.

\subsection{Median Learning Vector Quantization}

To facilitate fast metric learning, we would like to limit ourselves to as few positive
and negative pairs as possible. We propose to select positive and negative pairs via \emph{prototypical} data
points which represent the classes well. In particular, assume we have data points
$\tree_1, \ldots, \tree_\datalim$ with class assignments $\cls(\tree_1), \ldots, \cls(\tree_\datalim)$.
We propose to select a small sample of prototypes $\proto_1, \ldots, \proto_\protolim
\subset \{ \tree_1, \ldots, \tree_\datalim \}$ with $\protolim \ll \datalim$, and to construct
positive pairs for all $\tree_\dataidx$ as $(\tree_\dataidx, \proto_\dataidx^+)$,
where $\proto_\dataidx^+$ is the closest prototype to $\tree_\dataidx$ according to $\dist_{\cost_0}$
from the same class; and negative pairs as $(\tree_\dataidx, \proto_\dataidx^-)$, where
$\proto_\dataidx^-$ is the closest prototype to $\tree_\dataidx$ according to $\dist_{\cost_0}$
from a different class.

In our approach, we select these prototypical data points such that they help us to
discriminate between the classes. More precisely, we aim for prototypes $\proto_1, \ldots, \proto_\protolim$
which can classify as many data points correctly as possible by assigning the class
of the closest prototype. One way to obtain such prototypes is to optimize the 
generalized learning vector quantization (GLVQ) cost function \cite{Sato1995}:
\begin{equation}
\err = \sum_{\dataidx = 1}^\datalim \nonlin\Big(
\frac{\dist^+_\dataidx - \dist^-_\dataidx}{\dist^+_\dataidx + \dist^-_\dataidx}
\Big) \label{eq:glvq}
\end{equation}
where $\dist^+_\dataidx$ is the distance of $\tree_\dataidx$ to $\proto_\dataidx^+$,
$\dist^-_\dataidx$ is the distance of $\tree_\dataidx$ to $\proto_\dataidx^-$, and
$\nonlin(\mu) = \log(4 + \mu)$ \cite{Nebel2015}. Note that the fraction
$(\dist^+_\dataidx - \dist^-_\dataidx) / (\dist^+_\dataidx + \dist^-_\dataidx)$ is positive
if and only if $\tree_\dataidx$ is misclassified, such that the cost function is related to the
classification error.

Note that optimizing the GLVQ cost function in this case requires a discrete optimization scheme
because the prototypes $\proto_\protoidx$ are limited to be training data points, which is called
\emph{median} learning vector quantization \cite{Nebel2015}.
We follow the suggestion of \citet{Nebel2015} and apply a generalized expectation maximization (EM)
scheme to maximize $\sum_{\dataidx = 1}^\datalim \log(g^-_\dataidx + g^+_\dataidx)$, where
$g^-_\dataidx  =  2 + d^-_\dataidx / (d^+_\dataidx + d^-_\dataidx)$, and 
$g^+_\dataidx  =  2 - d^+_\dataidx / (d^+_\dataidx + d^-_\dataidx)$.
The expectation step of the EM scheme consists of computing the quantities
$\gamma^+_\dataidx = g^+_\dataidx / (g^+_\dataidx + g^-_\dataidx)$ as well as
$\gamma^-_\dataidx = g^-_\dataidx / (g^+_\dataidx + g^-_\dataidx)$ for all $\dataidx$,
and the maximization step consists of finding a prototype $\proto_\protoidx$ which can be set to a
different data point $\tree_\dataidx$ such that the likelihood $\likelihood
= \sum_{\dataidx = 1}^\datalim \gamma^+_\dataidx \cdot \log( g^+_\dataidx / \gamma^+_\dataidx)
+ \gamma^-_\dataidx \cdot \log( g^-_\dataidx / \gamma^-_\dataidx)$ is improved, assuming
fixed $\gamma^+_\dataidx$ and $\gamma^-_\dataidx$.
The EM scheme stops if it is not possible to improve $\likelihood$ for any prototype anymore.

\subsection{Metric Learning via Learning Vector Quantization}

Recall that the GLVQ cost function in Equation~\ref{eq:glvq} quantifies how well our
prototypes classify the training data. Following the recommendation of
\citet{Mokbel2015Neurocomputing}, we can not only use this cost function for learning
the prototypes, but also for learning the metric.
%
In particular, we can minimize the GLVQ loss with respect to the cost function $\cost$
using any unconstrained gradient-based solver, such as the limited-memory Broyden-Fletcher-Goldfarb-Shanno
(L-BFGS) algorithm \cite{Liu1989}. For the gradient $\grad{\cost} \err$ we obtain:
\begin{equation}
\sum_{\dataidx = 1}^\datalim \nonlin'\Big(\frac{\dist^+_\dataidx - \dist^-_\dataidx}{\dist^+_\dataidx + \dist^-_\dataidx}\Big)
\cdot \frac{\dist^-_\dataidx \cdot \grad{\cost} \dist^+_\dataidx - \dist^+_\dataidx \cdot \grad{\cost} \dist^-_\dataidx
}{(\dist^+_\dataidx + \dist^-_\dataidx)^2} \label{eq:grad_glvq}
\end{equation}
where $\nonlin'(\mu) = 1 / (4 + \mu)$.

Following the GESL approach of \citet{Bellet2012}, we optimize the pseudo tree edit distance $\tilde \dist_\cost$
instead of the tree edit distance itself, which yields the gradient
$\grad{\cost} \tilde \dist(\tree_\dataidx, \proto_\protoidx) = \freqmat_{\cost_0}(\tree_\dataidx, \proto_\protoidx)$.
However, we improve upon GESL by not only considering one cheapest edit script,
but instead the average over all cheapest edit scripts. In particular, we consider
$\freqmat_{\cost_0}(\tree_\dataidx, \proto_\protoidx)$ to be the average over the
matrices for all cheapest edit scripts.

Considering all co-optimal scripts permits us to exploit
additional information, with which we can prevent many degenerate cases in which $\tilde \dist_\cost$
underestimates $\dist_\cost$. In particular, the counter example in the proof of
Theorem~\ref{thm:gesl_degeneration} does not hold in this case. Computing this average over
all cheapest edit scripts is possible efficiently via a novel forward-backward algorithm
which we developed for this contribution (refer to the supplementary material;
\citet{Paassen2018Arxiv}).

We further note that changes to the metric may also enable us to optimize the prototype
locations further. Therefore, we employ an alternating optimization scheme where we first learn
the prototype positions according to median GLVQ, then adapt the metric, and repeat
until either the prototype positions do not change anymore or the solver is not able to improve
the metric anymore.

Until now, we have addressed the selection of positive and negative pairs, as well as a closer
coupling between pseudo edit distance and edit distance. However, we still have to ensure
pseudo-metric properties on the learned cost function. For this purpose, we introduce vectorial
embeddings.

\subsection{Tree Label Embeddings}

Let $\alphabet$ be a finite set with $\alphlim$ elements. Then, a vector embedding of $\alphabet$ is a
matrix $\embedmat \in \R^{\embeddim \times \alphlim}$ with $\embeddim \leq \alphlim$, where each
column is a vector $\embedvec(\node)$ for one $\node \in \alphabet$. Further, we define $\embedvec(\gap) := \vec 0$,
i.e.\ the origin of the $\embeddim$-dimensional Euclidean space.
We define the cost function corresponding to an embedding as the Euclidean distance between
the embedding vectors, that is:
$\cost_{\embedmat}(\lnode, \rnode) := \lVert \embedvec(\lnode) - \embedvec(\rnode) \rVert$.

Because the cost function is the Euclidean distance, it is guaranteed 
to be a pseudo-metric, irrespective of the choice of the embedding $\embedmat$.
Furthermore, $\cost_{\embedmat}$ is differentiable with respect to the embedding vectors with the gradient
$\grad{\embedvec(\lnode)} \cost_{\embedmat}(\lnode, \rnode) =
(\embedvec(\lnode) - \embedvec(\rnode)) / \lVert \embedvec(\lnode) - \embedvec(\rnode) \rVert$.
Using this gradient and Equation~\ref{eq:freqmat}, we can also obtain a gradient of
the pseudo-edit distance $\tilde \dist_\cost(\ltree, \rtree)$ with respect to $\embedvec(\lnode)$:
\begin{align}
&\grad{\embedvec(\lnode)} \tilde \dist_{\cost_{\embedmat}}(\ltree, \rtree) = \label{eq:grad_embed} \\
&\sum_{\lnodeidx = 1}^{\siz{\ltree}} \delta(\lnode, \lnode_\lnodeidx) \cdot
\bigg[\sum_{\rnodeidx = 1}^{\siz{\rtree} + 1} \freqmat_{\cost_0}(\ltree, \rtree)_{\lnodeidx, \rnodeidx} \cdot
\frac{\embedvec(\lnode) - \embedvec(\rnode_\rnodeidx)}{\lVert \embedvec(\lnode) - \embedvec(\rnode_\rnodeidx) \rVert} \bigg] \notag \\
+&\sum_{\rnodeidx = 1}^{\siz{\rtree}} \delta(\lnode, \rnode_\rnodeidx) \cdot \bigg[
\sum_{\lnodeidx = 1}^{\siz{\ltree} + 1} \freqmat_{\cost_0}(\ltree, \rtree)_{\lnodeidx, \rnodeidx} \cdot
\frac{\embedvec(\lnode) - \embedvec(\lnode_\lnodeidx)}{\lVert \embedvec(\lnode) - \embedvec(\lnode_\lnodeidx) \rVert} \bigg] \notag
\end{align}
where $\delta$ is the Kronecker-Delta, i.e.: $\delta(\lnode, \rnode) = 1$ if $\lnode = \rnode$ and
$0$ otherwise.

Finally, we can plug this result into Equation~\ref{eq:grad_glvq}, which yields a gradient
$\grad{\embedvec(\lnode)} \err$, such that we can learn the vectorial embedding of $\alphabet$
via gradient techniques. Note that prior theory on metric learning on the GLVQ cost function suggests
that the learned embedding will degenerate to a very low rank-matrix such that the model may
become overly simplistic \cite{Biehl2015}. To prevent such a degeneration,
we follow the regularization recommendation of \citet{Schneider2010} and add the term
$\regul \cdot \log(\det(\transp{\embedmat} \cdot \embedmat))$ to the GLVQ loss~\ref{eq:glvq},
which adds the gradient $\regul \cdot 2 \cdot \transp{{\embedmat^\dag}}$ where $\embedmat^\dag$ is the
Moore-Penrose-Pseudoinverse of $\embedmat$.
Additionally, we follow the regularization approach of good edit similarity learning \cite{Bellet2012}
and add the Frobenius-norm $\regul \cdot \lVert \embedmat \rVert^2_F$ to the loss, which
adds the gradient $\regul \cdot 2 \cdot \embedmat$.

As initialization of the vectorial embedding we use a $\alphlim$-dimensional simplex with
side length $1$, which leads to $\cost_0(\lnode, \rnode) = 0$ if $\lnode = \rnode$
and $1$ otherwise (refer to the supplementary material for a more detailed look into
this initialization \cite{Paassen2018ICMLSupp}).

Regarding computational complexity, we can analyze the gradient computation.
To compute a gradient, we first need to select the closest correct and closest wrong
prototype for every data point, which is possible in $\effic(\datalim \cdot \protolim)$.
Then, we need to compute the gradient for each data point via Equation~\ref{eq:grad_embed},
which is possible in $\effic(\datalim \cdot \rnodelim^2 \cdot \embeddim)$ where
$\rnodelim$ is the largest tree size in the data set. Computing the regularization
requires $\effic(\embeddim^3)$ due to the Pseudoinverse, resulting in
$\effic(\datalim \cdot \protolim + \datalim \cdot \rnodelim^2 \cdot \embeddim +
\embeddim^3)$ overall. How often the gradient needs to be computed depends on the
optimizer, but can typically be regarded as a constant. In our experiments, we limit
the number of gradient computations to $200$.

\section{Experiments}

In our experiment, we investigate whether our proposed metric learning scheme,
embedding edit distance learning (BEDL), is able to
improve classification accuracy beyond the default initialization, whether BEDL
improves upon the accuracy obtained by good edit similarity learning \cite{Bellet2012},
and whether the resulting embedding yields insight regarding the classification task in question.
In particular, we evaluate on the following data sets, including a variety of domains and
data set sizes.

\textbf{Strings:} A two-class data set of $200$ strings of length 12, adapted from~\citet{Mokbel2015Neurocomputing}.
Strings in class 1 consist of 6 $\sym{a}$ or $\sym{b}$ symbols, followed by a $\sym{c}$ or $\sym{d}$,
followed by another 5 $\sym{a}$ or $\sym{b}$ symbols. Which of the two respective symbols is
selected is chosen uniformly at random. Strings in class 2 are constructed in much the same way,
except that they consist of 5 $\sym{a}$ or $\sym{b}$ symbols, followed by a $\sym{c}$ or $\sym{d}$,
followed by another 6 $\sym{a}$ or $\sym{b}$ symbols. Note that the classes can be neither
discriminated via length nor via symbol frequency features. The decisive discriminative feature
is where a $\sym{c}$ or $\sym{d}$ is located in the string.

\textbf{MiniPalindrome} and \textbf{Sorting:} Two data sets of Java programs, where classes
represent different strategies to solve a programming task. The MiniPalindrome data set contains 48
programs implementing one of eight strategies to detect whether an input string contains only
palindromes \cite{MiniPalindrome}, and the Sorting data set contains 64 programs implementing either a BubbleSort or an InsertionSort strategy \cite{Sorting}.
The programs are represented by their abstract syntax tree where the label corresponds to one of
24 programming concepts (e.g. class declaration, function declaration, method call, etc.).

\textbf{Cystic} and \textbf{Leukemia:} Two data sets from KEGG/Glycan data base \cite{Glycan}
adapted from \citet{Gallicchio2013}, where one class corresponds to benign molecules and the other
class corresponds to molecules associated with cystic fibrosis or leukemia respectively.
The molecules are represented as trees, where the label corresponds to mono-saccharide identifiers
(one of 29 and one of 57 for Cystic and Leukemia, respectively), and the roots are chosen according
to biological meaning \cite{Glycan}. The cystic data set contains 160, the Leukemia data set 442 molecules.

\textbf{Sentiment:} A large-scale two-class data set of 9613 sentences from movie reviews,
where one class (4650 trees) corresponds to negative and the other class (4963 trees) to positive
reviews. The sentences are represented by their syntax trees, where inner nodes are unlabeled and
leaves are labeled with one of over $30,000$ words \cite{Sentiment}.
Note that GESL is not practically applicable for this data set, as the number of parameters to
learn scales quadratically with the number of words, i.e.\ $> 30,000^2$. To make BEDL applicable
in whis case, we initialize the vectorial embedding with the $300$-dimensional Common Crawl GloVe embedding
\cite{Glove}, which we reduce via PCA, retaining $95\%$ of the data variance ($\embeddim = 16.4 \pm 2.3$ dimensions on
average $\pm$ standard deviation).
We adapt this initial embedding via a linear transformation $\relproj \in \R^{\embeddim \times \embeddim}$
which we learn vie BEDL.
Further, we replace the cost function with the cosine distance
$\cost_\relproj(\vec \lnode, \vec \rnode) = \frac{1}{2} - \frac{1}{2} \cdot (\transp{(\relproj \cdot \vec \lnode)} \cdot \relproj \cdot \vec \rnode)
/ (\lVert \relproj \cdot \vec \lnode \rVert \cdot \lVert \relproj \cdot \vec \rnode \rVert)$, which is the
recommended distance measure for the GloVe word embedding \cite{Glove}
(refer to the supplementary material for the gradient; \citet{Paassen2018ICMLSupp}).

On each data set, we perform a crossvalidation\footnote{We used 20 folds for Strings and Sentiment,
10 for Cystic and Leukemia, 8 for Sorting and 6 for MiniPalindrome. For the programming data sets,
the number of folds had to be reduced to ensure that each fold still contained a meaningful
number of data points. For the Cystic and Leukemia data set, our ten folds were consistent
with the paper of \citet{Gallicchio2013}. In all cases, folds were generated such that
the label distribution of the overall data set was maintained.} and compare the average test error across folds.
In particular, we compare the error when using the initial tree edit distance with the error when using the
pseudo-edit distance learned via good edit similarity learning (GESL), and the tree edit distance
learned via our proposed approach (BEDL).

In general, we would expect that a discriminative metric learned for one classifier also facilitates
classification using other classifiers. Therefore, we report the classification error for four
classifiers, namely the median generalized learning vector quantization classifier (MGLVQ)
for which our method is optimized, the goodness classifier for which GESL is optimized \cite{Bellet2012},
the K-nearest neighbor (KNN) classifier, and the support vector machine (SVM) based on the radial
basis function kernel. In order to ensure a kernel matrix for SVM, we set negative eigenvalues to
zero (clip Eigenvalue correction). Note that this eigenvalue
correction requires cubic runtime in terms of the number of data points and is thus prohibitively
slow for large data set sizes. Therefore, for the Sentiment data set,
we trained the classifiers on a randomly selected sample of $300$ points from the training data.

We optimized all hyper-parameters in a nested $5$-fold crossvalidation, namely
the number of prototypes $\protolim$ for MGLVQ and LVQ metric learning in the range $[1, 15]$,
the number of neighbors for KNN in the range $[1, 15]$, the kernel bandwidth for SVM in the range $[0.1,10]$,
the sparsity parameter $\sparsity$ for the goodness classifier in the range $[10^-5, 10]$, and the
regularization strength $\regul$ for GESL and BEDL in the range $2 \cdot \protolim \cdot \datalim \cdot [10^{-6}, 10^{-2}]$.
We chose the number of prototypes for BEDL, as well as the number of
neighbors for GESL as the optimal number of prototypes $\protolim$ for MGLVQ.

As implementations, we used custom implementations of KNN,
MGLVQ, the goodness classifier, GESL, and BEDL, which are availabe at
\url{https://doi.org/10.4119/unibi/2919994}.
For SVM, we utilized the LIBSVM standard implementation \cite{Chang2011}.
All experiments were performed on a consumer-grade laptop with an Intel Core i7-7700 HQ CPU.

\begin{table}
\label{tab:results}
\caption{The mean test classification error and runtimes for metric learning,
averaged over the cross validation trials, as well as the standard deviation.
The x-axis shows the metric learning schemes, the y-axis the different classifiers
used for evaluation. The table is sub-divided for each data set.
The lowest classification error for each data set is highlighted via bold print.}
\vspace{0.3cm}
\begin{scriptsize}
\begin{tabular}{lccc}
classifier & initial & GESL & BEDL \\
\cmidrule(lr){1-1} \cmidrule(lr){2-4}
& & Strings & \\
\cmidrule(lr){1-1} \cmidrule(lr){2-4}
KNN          & $21.0 \pm 10.2 \%$ & $23.0 \pm 10.8 \%$ & $\bm{0.0 \pm 0.0 \%}$ \\
MGLVQ       & $36.0 \pm 15.7 \%$ & $34.0 \pm 11.0 \%$ & $\bm{0.0 \pm 0.0 \%}$ \\
SVM          & $9.0 \pm 11.2 \%$  & $10.0 \pm 8.6 \%$  & $\bm{0.0 \pm 0.0 \%}$ \\
goodness     & $11.5 \pm 9.3 \%$  & $0.5 \pm 2.2 \%$   & $\bm{0.0 \pm 0.0 \%}$ \\
runtime [s]  & $0 \pm 0 $         & $0.030 \pm 0.002 $ & $1.077 \pm 0.098$ \\
\cmidrule(lr){1-1} \cmidrule(lr){2-4}
& & MiniPalindrome & \\
\cmidrule(lr){1-1} \cmidrule(lr){2-4}
KNN          & $12.5 \pm 11.2 \%$ & $12.5 \pm 7.9 \%$  & $10.4 \pm 9.4 \%$ \\
MGLVQ       & $2.1 \pm 5.1 \%$   & $4.2 \pm 6.5 \%$   & $\bm{0.0 \pm 0.0 \%}$ \\
SVM          & $4.2 \pm 6.5 \%$   & $20.8 \pm 15.1 \%$ & $\bm{0.0 \pm 0.0 \%}$ \\
goodness     & $6.2 \pm 6.8 \%$   & $14.6 \pm 5.1 \%$  & $8.3 \pm 10.2 \%$ \\
runtime [s]  & $0 \pm 0$          & $0.103 \pm 0.014$  & $2.785 \pm 0.631$ \\
\cmidrule(lr){1-1} \cmidrule(lr){2-4}
& & Sorting & \\
\cmidrule(lr){1-1} \cmidrule(lr){2-4}
KNN          & $15.6 \pm 8.8 \%$  & $18.8 \pm 16.4 \%$ & $10.9 \pm 8.0 \%$ \\
MGLVQ       & $14.1 \pm 10.4 \%$ & $14.1 \pm 8.0 \%$  & $14.1 \pm 8.0 \%$ \\
SVM          & $10.9 \pm 8.0 \%$  & $\bm{9.4 \pm 8.8 \%}$ & $\bm{9.4 \pm 8.8 \%}$ \\
goodness     & $15.6 \pm 11.1 \%$ & $17.2 \pm 14.8 \%$ & $17.2 \pm 9.3 \%$ \\
runtime [s]  & $0 \pm 0$          & $0.352 \pm 0.102$  & $3.358 \pm 0.748$ \\
\cmidrule(lr){1-1} \cmidrule(lr){2-4}
& & Cystic & \\
\cmidrule(lr){1-1} \cmidrule(lr){2-4}
KNN          & $31.2 \pm 6.6 \%$ & $32.5 \pm 10.1 \%$ & $28.1 \pm 8.5 \%$ \\
MGLVQ       & $34.4 \pm 6.8 \%$ & $33.1 \pm 9.8 \%$  & $30.0 \pm 10.1 \%$ \\
SVM          & $28.1 \pm 9.0 \%$ & $33.1 \pm 8.9 \%$  & $29.4 \pm 12.5 \%$ \\
goodness     & $28.1 \pm 8.5 \%$ & $26.2 \pm 14.4 \%$ & $\bm{24.4 \pm 13.3 \%}$ \\
runtime [s]  & $0 \pm 0$         & $0.353 \pm 0.292$  & $0.864 \pm 0.767$ \\
\cmidrule(lr){1-1} \cmidrule(lr){2-4}
& & Leukemia & \\
\cmidrule(lr){1-1} \cmidrule(lr){2-4}
KNN          & $7.5 \pm 2.6 \%$ & $8.2 \pm 4.6 \%$  & $7.3 \pm 4.3 \%$ \\
MGLVQ       & $9.5 \pm 4.0 \%$ & $10.9 \pm 4.7 \%$ & $9.5 \pm 3.0 \%$ \\
SVM          & $7.0 \pm 4.1 \%$ & $8.8 \pm 2.9 \%$  & $6.8 \pm 4.7 \%$ \\
goodness     & $\bm{6.1 \pm 4.3 \%}$ & $10.0 \pm 4.4 \%$ & $6.3 \pm 3.8 \%$ \\
runtime [s]  & $0 \pm 0$        & $2.208 \pm 0.919$ & $6.550 \pm 2.706$ \\
\cmidrule(lr){1-1} \cmidrule(lr){2-4}
& & Sentiment & \\
\cmidrule(lr){1-1} \cmidrule(lr){2-4}
kNN          & $40.2 \pm 2.8 \%$ & $-$              & $38.2 \pm 3.3 \%$ \\
MGLVQ       & $44.0 \pm 2.6 \%$ & $-$              & $41.3 \pm 5.7 \%$ \\
SVM          & $34.3 \pm 3.0 \%$ & $-$              & $\bm{33.3 \pm 3.6 \%}$ \\
goodness     & $43.7 \pm 1.9 \%$ & $-$              & $42.5 \pm 3.1 \%$ \\
runtime [s]  & $0 \pm 0$         & $-$              & $69.385 \pm 58.064$ \\
\end{tabular}
\end{scriptsize}
\end{table}
\pgfplotsset{
	embedding/.style={%
		only marks,
		draw=skyblue3, skyblue3, thick,
		fill=skyblue1,
		mark=*
	}
}
\begin{figure}
\begin{center}
\begin{tikzpicture}
\begin{axis}[disabledatascaling, 
width=0.4\textwidth, height=2.5cm,
ytick={-0.1, 0, 0.1}, xmin=0, xmax=4, ymin=-0.1, ymax=0.1, enlarge x limits]
\addplot[embedding] table[x=x, y=y] {string_embedding.csv};
\node[above right, skyblue3] at (0,0) {$\sym{a}, \sym{b}, \gap$};
\node[above, skyblue3] at (3.751,0) {$\sym{c}, \sym{d}$};
\end{axis}
\end{tikzpicture}

\hspace{0.2cm}
\begin{tikzpicture}
\begin{axis}[view={60}{30}, disabledatascaling, 
width=0.4\textwidth, height=0.25\textwidth,
xmin=0, xmax=10, ymin=-5, ymax=+5, zmin=-5, zmax=+2, enlarge x limits, enlarge y limits]
\addplot3[embedding] table[x=x, y=y, z=z2, col sep=comma] {mini_palindrome_embedding.csv};
\node[above right, skyblue3] at (0,0) {$\gap$};
\node[left, skyblue3] at (6,-4.5) {block};
\node[below right, skyblue3] at (9,0.2) {modifiers};
\node[below right, skyblue3] at (5,5.6) {while};
\node[below, outer sep=1cm, skyblue3] at (1.7,0) {parameterized type};
\end{axis}
\end{tikzpicture}
\end{center}
\caption{A PCA of the learned embeddings for the Strings (top) and the
MiniPalindrome data set (bottom), covering $100\%$ and $83.54\%$ of the variance respectively.}
\label{fig:embeddings}
\end{figure}
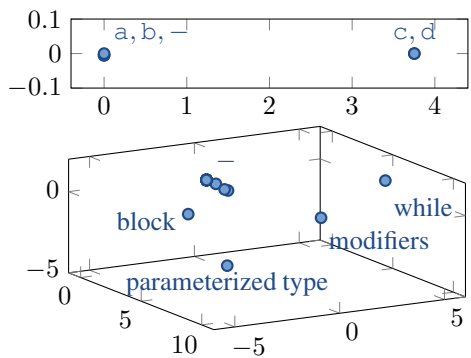

The results of our experiments are displayed in Table~\ref{tab:results}. In all data sets
and for all classifiers, BEDL yields lower classification error compared to GESL.
For the Strings data set we can also verify this result statistically with a one-sided Wilcoxon
signed rank test ($p < 10^{-4}$).
Furthermore, in all but the Leukemia data set, BEDL yields the overall best classification
results, and is close to optimal for the Leukemia data set ($0.2\%$ difference).
In all cases, BEDL could improve the accuracy for KNN, in five out of six cases for
SVM (the exception being the Cystic data set), in four out of six cases for MGLVQ (in Sorting and
Leukemia it stayed equal), and in three out of six cases for the goodness classifier.
For the Strings and Sentiment data sets we can also verify this result statistically with
$p < 0.05$ for all classifiers.

Note that the focus of our work is to improve classification accuracy via metric learning, not
to develop state-of-the-art classifiers as such. However, we note that our results for the Sorting
data set outperform the best reported results by \citet{Paassen2016Neurocomputing} of $15\%$. For the
Cystic data set we improve the AUC from $76.93 \pm 0.97 \%$ mean and standard deviation across
crossvalidation trials to $79.2 \pm 13.6 \%$, and for the Leukemia data set from
$93.8 \pm 3.3\%$ to $94.6 \pm 4.5\%$. Both values are competitive with the results obtained
via recursive neural networks and a multitude of graph kernels by \citet{Gallicchio2013}.
For the Sentiment data set, we obtain a SVM classification error of $27.51\%$ on the validation set, which is noticeably
worse than the reported literature results of around $12.5\%$ \cite{Sentiment}. However, we note that we
used considerably less data to train our classifier (only $500$ points for the validation).

Interestingly, GESL tended to decrease classification accuracy compared to the initial tree
edit distance. Likely, GESL requires more neighbors for better results \cite{Bellet2012}.
However, scaling up to a high number of neighbors lead to prohibitively high runtimes
for our experiments such that we do not report these results here. These high runtimes can
be explained by the fact that the number of slack variables in GESL increases with
$\effic(\datalim \cdot \protolim)$ where $\datalim$ is the number of data points and
$\protolim$ is the number of neighbors.
The scaling behavior is also visible in our experimental results. For data sets with little
data points and neighbors, such as Strings, MiniPalindrome, and Sorting, GESL is $10$ to
$30$ times faster compared to BEDL. However, for Cystic and Leukemia,
the runtime advantage shrinks to a factor of $2$ to $3$.

In ablation studies, we studied the difference between GESL and BEDL in
more detail. We observed that considering the average over all co-optimal edit scripts,
and considering LVQ prototypes instead of ad-hoc nearest neighbors, improved GESL
on the MiniPalindrome data set, worsened it for the Strings data set, and
showed no remarkable difference for the Sorting, Cystic, and Leukemia data set.
We also compared BEDL without the embedding approach and with the
embedding approach. Interestingly, the pseudo-edit distance performed worse when
considering embeddings, while the actual edit distance performed better when
considering embeddings. In general, GESL variants performed better for the
pseudo-edit distance than for the actual edit distance, and LVQ variants performed
better for the actual edit distance compared to the pseudo edit distance.
We report the full results in the supplementary material \cite{Paassen2018ICMLSupp}.

Beyond classification accuracy, our metric learning approach permits to inspect the
resulting embedding. Figure~\ref{fig:embeddings} displays a PCA of the embeddings learned
for the Strings and MiniPalindrome data set respectively. As we can see, the embedding
for the Strings data set captures the objective of the task perfectly: Both $\sym{a}$ as
well as $\sym{b}$ symbols are irrelevant for class discrimination and are thus embedded
at the origin, while $\sym{c}$ and $\sym{d}$ are embedded far from the origin, but both
at the same location. For MiniPalindrome, we also observe that most syntactic concepts are
embedded at zero, indicating that a combination of the four remaining concepts is sufficient
for class discrimination; namely the block concept, which captures the nesting structure of the program,
the while concept, which is specific to one of the classes, the modifiers concept, which can
serve to count the number of variables and methods in the program, and the parametrized
type concept, which distinguishes programs with advanced data structured from programs with
primitive data structures.

\section{Conclusion}

In this contribution we have proposed embedding edit distance learning (BEDL) as a novel approach
for edit distance learning on trees with three distinct characteristics:
First, our objective is the generalized learning vector quantization
cost function, which pulls data points closer to the closest prototype for their own class and
pushes them away from the closest prototype for a different class; second, we consider not only
a single optimal edit script between trees but a summary of \emph{all} co-optimal edit scripts;
finally, we do not learn a cost function for the edits directly, but instead a vectorial embedding
of the label alphabet, which guarantees metric properties and can be interpreted.
In our experiments we have shown that BEDL improves upon the
state-of-the-art of good edit similarity learning for trees on a diverse tree data sets
including Java program syntax trees, tree-based molecule representations from a biomedical task,
and syntax trees in natural language processing.

Limitations of our work include that an improvement of the loss function for the pseudo
edit distance does not strictly imply an improvement of the loss for the actual
edit distance, and that improvements in classification accuracy are small for some
classifiers and some data sets.
Future research should investigate the relation between pseudo edit distance and edit distance,
as well as the relation between the number of prototypes and metric learning performance in
more detail. It may also be worthwhile to study different cost functions, in particular probabilistic
ones, which may be compatible with probabilistic models of the edit distance. Still, we regard our
existing contribution as a meaningful step towards edit distances on trees which are both
discriminative as well as interpretable and can thus enhance accuracy and understanding on
classification tasks of structured data.

\clearpage

\section*{Acknowledgements}

Funding by the DFG under grant number HA 2719/6-2 and the CITEC center of 
excellence (EXC 277) is gratefully acknowledged.
Thanks to David Nebel and our anonymous reviewers for helpful comments
and suggestions.

\bibliography{literature}

\begin{thebibliography}{41}
\providecommand{\natexlab}[1]{#1}
\providecommand{\url}[1]{\texttt{#1}}
\expandafter\ifx\csname urlstyle\endcsname\relax
  \providecommand{\doi}[1]{doi: #1}\else
  \providecommand{\doi}{doi: \begingroup \urlstyle{rm}\Url}\fi

\bibitem[Aiolli \& Donini(2015)Aiolli and Donini]{Aiolli2015MKL}
Aiolli, F. and Donini, M.
\newblock Easy{MKL}: {A} scalable multiple kernel learning algorithm.
\newblock \emph{Neurocomputing}, 169:\penalty0 215--224, 2015.
\newblock \doi{10.1016/j.neucom.2014.11.078}.

\bibitem[Aiolli et~al.(2015)Aiolli, Martino, and Sperduti]{Aiolli2015}
Aiolli, F., Martino, G. D.~S., and Sperduti, A.
\newblock An efficient topological distance-based tree kernel.
\newblock \emph{{IEEE} Transactions on Neural Networks and Learning Systems},
  26\penalty0 (5):\penalty0 1115--1120, 2015.
\newblock \doi{10.1109/TNNLS.2014.2329331}.

\bibitem[Bacciu et~al.(2018)Bacciu, Micheli, and Sperduti]{Bacciu2018}
Bacciu, D., Micheli, A., and Sperduti, A.
\newblock Generative kernels for tree-structured data.
\newblock \emph{IEEE Transactions on Neural Networks and Learning Systems},
  2018.
\newblock \doi{10.1109/TNNLS.2017.2785292}.
\newblock in press.

\bibitem[Balcan et~al.(2008)Balcan, Blum, and Srebro]{Balcan2008}
Balcan, M.-F., Blum, A., and Srebro, N.
\newblock A theory of learning with similarity functions.
\newblock \emph{Machine Learning}, 72\penalty0 (1):\penalty0 89--112, Aug 2008.
\newblock \doi{10.1007/s10994-008-5059-5}.

\bibitem[Bellet et~al.(2012)Bellet, Habrard, and Sebban]{Bellet2012}
Bellet, A., Habrard, A., and Sebban, M.
\newblock Good edit similarity learning by loss minimization.
\newblock \emph{Machine Learning}, 89\penalty0 (1):\penalty0 5--35, Oct 2012.
\newblock \doi{10.1007/s10994-012-5293-8}.

\bibitem[Bellet et~al.(2014)Bellet, Habrard, and Sebban]{Bellet2014}
Bellet, A., Habrard, A., and Sebban, M.
\newblock A survey on metric learning for feature vectors and structured data.
\newblock \emph{arXiv e-prints}, 2014.
\newblock URL \url{http://arxiv.org/abs/1306.6709}.

\bibitem[Bellet et~al.(2016)Bellet, Bernabeu, Habrard, and Sebban]{Bellet2016}
Bellet, A., Bernabeu, J.~F., Habrard, A., and Sebban, M.
\newblock Learning discriminative tree edit similarities for linear
  classification - application to melody recognition.
\newblock \emph{Neurocomputing}, 214:\penalty0 155 -- 161, 2016.
\newblock \doi{10.1016/j.neucom.2016.06.006}.

\bibitem[Biehl et~al.(2015)Biehl, Hammer, Schleif, Schneider, and
  Villmann]{Biehl2015}
Biehl, M., Hammer, B., Schleif, F.-M., Schneider, P., and Villmann, T.
\newblock Stationarity of matrix relevance {LVQ}.
\newblock In \emph{Proceedings of the 2015 International Joint Confgerence on
  Neural Networks (IJCNN 2015)}, pp.\  1--8, 2015.
\newblock \doi{10.1109/IJCNN.2015.7280441}.

\bibitem[Boyer et~al.(2007)Boyer, Habrard, and Sebban]{Boyer2007}
Boyer, L., Habrard, A., and Sebban, M.
\newblock Learning metrics between tree structured data: {A}pplication to image
  recognition.
\newblock In Kok, J.~N., Koronacki, J., Mantaras, R. L.~d., Matwin, S.,
  Mladeni{\v{c}}, D., and Skowron, A. (eds.), \emph{Proceedings of the 18th
  European Conference on Machine Learning (ECML 2007)}, pp.\  54--66. Springer,
  2007.
\newblock \doi{10.1007/978-3-540-74958-5_9}.

\bibitem[Chang \& Lin(2011)Chang and Lin]{Chang2011}
Chang, C.-C. and Lin, C.-J.
\newblock {LIBSVM}: A library for support vector machines.
\newblock \emph{ACM Transactions on Intelligent Systems and Technology},
  2\penalty0 (3):\penalty0 27:1--27:27, 2011.
\newblock \doi{10.1145/1961189.1961199}.
\newblock Software available at \url{http://www.csie.ntu.edu.tw/~cjlin/libsvm}.

\bibitem[Cho et~al.(2014)Cho, van Merrienboer, G{\"{u}}l{\c{c}}ehre, Bougares,
  Schwenk, and Bengio]{Ch0214}
Cho, K., van Merrienboer, B., G{\"{u}}l{\c{c}}ehre, {\c{C}}., Bougares, F.,
  Schwenk, H., and Bengio, Y.
\newblock Learning phrase representations using {RNN} encoder-decoder for
  statistical machine translation.
\newblock In Moschitti, A., Pang, B., and Daelemans, W. (eds.),
  \emph{Proceedings of the 2014 Conference on Empirical Methods in Natural
  Language Processing (EMNLP 2014)}, pp.\  1724--1734, 2014.
\newblock URL \url{https://www.aclweb.org/anthology/D14-1179}.

\bibitem[Cover \& Hart(1967)Cover and Hart]{Cover1967}
Cover, T.~M. and Hart, P.~E.
\newblock Nearest neighbor pattern classification.
\newblock \emph{IEEE Transactions on Information Theory}, 13\penalty0
  (1):\penalty0 21--27, 1967.
\newblock \doi{10.1109/TIT.1967.1053964}.

\bibitem[Da~San~Martino \& Sperduti(2010)Da~San~Martino and
  Sperduti]{DaSanMartino2010}
Da~San~Martino, G. and Sperduti, A.
\newblock Mining structured data.
\newblock \emph{Computational Intelligence Magazine}, 5\penalty0 (1):\penalty0
  42--49, Feb 2010.
\newblock \doi{10.1109/MCI.2009.935308}.

\bibitem[Gallicchio \& Micheli(2013)Gallicchio and Micheli]{Gallicchio2013}
Gallicchio, C. and Micheli, A.
\newblock Tree echo state networks.
\newblock \emph{Neurocomputing}, 101:\penalty0 319--337, 2013.
\newblock \doi{10.1016/j.neucom.2012.08.017}.

\bibitem[Gisbrecht et~al.(2015)Gisbrecht, Schulz, and Hammer]{Gisbrecht2015}
Gisbrecht, A., Schulz, A., and Hammer, B.
\newblock Parametric nonlinear dimensionality reduction using kernel t-{SNE}.
\newblock \emph{Neurocomputing}, 147:\penalty0 71--82, 2015.
\newblock \doi{10.1016/j.neucom.2013.11.045}.

\bibitem[Hammer et~al.(2007)Hammer, Micheli, and Sperduti]{Hammer2007}
Hammer, B., Micheli, A., and Sperduti, A.
\newblock \emph{Adaptive Contextual Processing of Structured Data by Recursive
  Neural Networks: {A} Survey of Computational Properties}, pp.\  67--94.
\newblock Springer Berlin Heidelberg, Berlin, Heidelberg, 2007.
\newblock \doi{10.1007/978-3-540-73954-8_4}.

\bibitem[Hashimoto et~al.(2006)Hashimoto, Goto, Kawano, Aoki-Kinoshita, Ueda,
  Hamajima, Kawasaki, and Kanehisa]{Glycan}
Hashimoto, K., Goto, S., Kawano, S., Aoki-Kinoshita, K.~F., Ueda, N., Hamajima,
  M., Kawasaki, T., and Kanehisa, M.
\newblock {{K}{E}{G}{G} as a glycome informatics resource}.
\newblock \emph{Glycobiology}, 16\penalty0 (5):\penalty0 63R--70R, 2006.
\newblock \doi{10.1093/glycob/cwj010}.

\bibitem[Kohonen(1995)]{Kohonen1995}
Kohonen, T.
\newblock \emph{Learning Vector Quantization}, pp.\  175--189.
\newblock Springer Berlin Heidelberg, 1995.
\newblock \doi{10.1007/978-3-642-97610-0_6}.

\bibitem[Kulis(2013)]{Kulis2013}
Kulis, B.
\newblock Metric learning: {A} survey.
\newblock \emph{Foundations and Trends in Machine Learning}, 5\penalty0
  (4):\penalty0 287--364, 2013.
\newblock \doi{10.1561/2200000019}.

\bibitem[Liu \& Nocedal(1989)Liu and Nocedal]{Liu1989}
Liu, D.~C. and Nocedal, J.
\newblock On the limited memory {BFGS} method for large scale optimization.
\newblock \emph{Mathematical Programming}, 45\penalty0 (1):\penalty0 503--528,
  1989.
\newblock \doi{10.1007/BF01589116}.

\bibitem[Mokbel et~al.(2015)Mokbel, Paa{\ss}en, Schleif, and
  Hammer]{Mokbel2015Neurocomputing}
Mokbel, B., Paa{\ss}en, B., Schleif, F.-M., and Hammer, B.
\newblock Metric learning for sequences in relational {LVQ}.
\newblock \emph{Neurocomputing}, 169:\penalty0 306--322, 2015.
\newblock \doi{10.1016/j.neucom.2014.11.082}.

\bibitem[Nebel et~al.(2015)Nebel, Hammer, Frohberg, and Villmann]{Nebel2015}
Nebel, D., Hammer, B., Frohberg, K., and Villmann, T.
\newblock Median variants of learning vector quantization for learning of
  dissimilarity data.
\newblock \emph{Neurocomputing}, 169:\penalty0 295--305, 2015.
\newblock \doi{10.1016/j.neucom.2014.12.096}.

\bibitem[Paa{\ss}en et~al.(2016)Paa{\ss}en, Mokbel, and
  Hammer]{Paassen2016Neurocomputing}
Paa{\ss}en, B., Mokbel, B., and Hammer, B.
\newblock Adaptive structure metrics for automated feedback provision in
  intelligent tutoring systems.
\newblock \emph{Neurocomputing}, 192:\penalty0 3--13, 2016.
\newblock \doi{10.1016/j.neucom.2015.12.108}.

\bibitem[Paa{\ss}en et~al.(2018)Paa{\ss}en, Hammer, Price, Barnes, Gross, and
  Pinkwart]{Paassen2018JEDM}
Paa{\ss}en, B., Hammer, B., Price, T.~W., Barnes, T., Gross, S., and Pinkwart,
  N.
\newblock The {Continuous Hint Factory} - providing hints in vast and sparsely
  populated edit distance spaces.
\newblock \emph{Journal of Educational Datamining}, 2018.
\newblock URL \url{http://arxiv.org/abs/1708.06564}.
\newblock accepted.

\bibitem[Paaßen(2016{\natexlab{a}})]{MiniPalindrome}
Paaßen, B.
\newblock {MiniPalindrome}, 2016{\natexlab{a}}.
\newblock Bielefeld University, \doi{10.4119/unibi/2900666}.

\bibitem[Paaßen(2016{\natexlab{b}})]{Sorting}
Paaßen, B.
\newblock {Java Sorting Programs}, 2016{\natexlab{b}}.
\newblock Bielefeld University, \doi{10.4119/unibi/2900684}.

\bibitem[Paaßen(2018{\natexlab{a}})]{Paassen2018Arxiv}
Paaßen, B.
\newblock Revisiting the tree edit distance and its backtracing: {A} tutorial.
\newblock \emph{ArXiv e-prints}, 2018{\natexlab{a}}.
\newblock URL \url{https://arxiv.org/abs/1805.06869}.

\bibitem[Paaßen(2018{\natexlab{b}})]{Paassen2018ICMLSupp}
Paaßen, B.
\newblock Tree edit distance learning via adaptive symbol embeddings:
  {S}upplementary materials and results.
\newblock \emph{Ar{X}iv e-prints}, 2018{\natexlab{b}}.
\newblock URL \url{https://arxiv.org/abs/1805.07123}.

\bibitem[Pennington et~al.(2014)Pennington, Socher, and Manning]{Glove}
Pennington, J., Socher, R., and Manning, C.~D.
\newblock {GloVe}: Global vectors for word representation.
\newblock In Moschitti, A., Pang, B., and Daelemans, W. (eds.),
  \emph{Proceedings of the 2014 Conference on Empirical Methods in Natural
  Language Processing (EMNLP 2014)}, pp.\  1532--1543, 2014.
\newblock URL \url{http://www.aclweb.org/anthology/D14-1162}.

\bibitem[Sato \& Yamada(1995)Sato and Yamada]{Sato1995}
Sato, A. and Yamada, K.
\newblock Generalized learning vector quantization.
\newblock In Tesauro, G., Touretzky, D., and Leen, T. (eds.), \emph{Proceedings
  of the 7th conference on Advances in Neural Information Processing Systems
  (NIPS 1995)}, pp.\  423--429. MIT Press, 1995.
\newblock URL
  \url{https://papers.nips.cc/paper/1113-generalized-learning-vector-quantization}.

\bibitem[Schleif \& Tino(2015)Schleif and Tino]{Schleif2015}
Schleif, F.-M. and Tino, P.
\newblock Indefinite proximity learning: {A} review.
\newblock \emph{Neural Computation}, 27\penalty0 (10):\penalty0 2039--2096,
  2015.
\newblock \doi{10.1162/NECO_a_00770}.

\bibitem[Schneider et~al.(2009)Schneider, Biehl, and Hammer]{Schneider2009}
Schneider, P., Biehl, M., and Hammer, B.
\newblock Adaptive relevance matrices in learning vector quantization.
\newblock \emph{Neural Computation}, 21\penalty0 (12):\penalty0 3532--3561,
  2009.
\newblock \doi{10.1162/neco.2009.11-08-908}.

\bibitem[Schneider et~al.(2010)Schneider, Bunte, Stiekema, Hammer, Villmann,
  and Biehl]{Schneider2010}
Schneider, P., Bunte, K., Stiekema, H., Hammer, B., Villmann, T., and Biehl, M.
\newblock Regularization in matrix relevance learning.
\newblock \emph{IEEE Transactions on Neural Networks}, 21\penalty0
  (5):\penalty0 831--840, 2010.
\newblock \doi{10.1109/TNN.2010.2042729}.

\bibitem[Shervashidze et~al.(2011)Shervashidze, Schweitzer, Leeuwen, Mehlhorn,
  and Borgwardt]{Shervashidze2011}
Shervashidze, N., Schweitzer, P., Leeuwen, E. J.~v., Mehlhorn, K., and
  Borgwardt, K.~M.
\newblock Weisfeiler-{L}ehman graph kernels.
\newblock \emph{Journal of Machine Learning Research}, 12\penalty0
  (Sep):\penalty0 2539--2561, 2011.
\newblock URL \url{http://www.jmlr.org/papers/v12/shervashidze11a.html}.

\bibitem[Smith \& Waterman(1981)Smith and Waterman]{Smith1981}
Smith, T.~F. and Waterman, M.~S.
\newblock Identification of common molecular subsequences.
\newblock \emph{Journal of Molecular Biology}, 147\penalty0 (1):\penalty0
  195--197, 1981.
\newblock \doi{10.1016/0022-2836(81)90087-5}.

\bibitem[Socher et~al.(2013)Socher, Perelygin, Wu, Chuang, Manning, Ng, and
  Potts]{Sentiment}
Socher, R., Perelygin, A., Wu, J.~Y., Chuang, J., Manning, C.~D., Ng, A.~Y.,
  and Potts, C.
\newblock Recursive deep models for semantic compositionality over a sentiment
  treebank.
\newblock In Baldwin, T. and Korhonen, A. (eds.), \emph{Proceedings of the 2013
  Conference on Empirical Methods in Natural Language Processing (EMNLP 2013)},
  pp.\  1631--1642, 2013.
\newblock URL \url{http://www.aclweb.org/anthology/D/D13/D13-1170.pdf}.

\bibitem[Sutskever et~al.(2014)Sutskever, Vinyals, and Le]{Sutskever2014}
Sutskever, I., Vinyals, O., and Le, Q.~V.
\newblock Sequence to sequence learning with neural networks.
\newblock In Ghahramani, Z., Welling, M., Cortes, C., Lawrence, N.~D., and
  Weinberger, K.~Q. (eds.), \emph{Proceedings of the 27th Conference on
  Advances in Neural Information Processing Systems (NIPS 2014)}, pp.\
  3104--3112. Curran Associates, Inc., 2014.
\newblock URL
  \url{https://papers.nips.cc/paper/5346-sequence-to-sequence-learning-with-neural-networks}.

\bibitem[Yanardag \& Vishwanathan(2015)Yanardag and Vishwanathan]{Yanardag2015}
Yanardag, P. and Vishwanathan, S.
\newblock Deep graph kernels.
\newblock In \emph{Proceedings of the 21th International Conference on
  Knowledge Discovery and Data Mining (KDD 2015)}, pp.\  1365--1374, New York,
  NY, USA, 2015. ACM.
\newblock \doi{10.1145/2783258.2783417}.

\bibitem[Zeng et~al.(2009)Zeng, Tung, Wang, Feng, and Zhou]{Zeng2009}
Zeng, Z., Tung, A. K.~H., Wang, J., Feng, J., and Zhou, L.
\newblock Comparing stars: {O}n approximating graph edit distance.
\newblock \emph{Proceedings of the VLDB Endowment}, 2\penalty0 (1):\penalty0
  25--36, 2009.
\newblock \doi{10.14778/1687627.1687631}.

\bibitem[Zhang \& Shasha(1989)Zhang and Shasha]{Zhang1989}
Zhang, K. and Shasha, D.
\newblock Simple fast algorithms for the editing distance between trees and
  related problems.
\newblock \emph{SIAM Journal on Computing}, 18\penalty0 (6):\penalty0
  1245--1262, 1989.
\newblock \doi{10.1137/0218082}.

\bibitem[Zhang et~al.(1992)Zhang, Statman, and Shasha]{Zhang1992}
Zhang, K., Statman, R., and Shasha, D.
\newblock On the editing distance between unordered labeled trees.
\newblock \emph{Information Processing Letters}, 42\penalty0 (3):\penalty0
  133--139, 1992.
\newblock \doi{10.1016/0020-0190(92)90136-J}.

\end{thebibliography}
\bibliographystyle{icml2018}

\end{document}